\documentclass[review,authoryear]{elsarticle}
\usepackage{natbib}
\usepackage{subfigure}
\usepackage{lineno}
\usepackage[colorlinks,urlcolor=blue]{hyperref}
\usepackage{amsmath,amssymb}
\usepackage{diagbox}
\usepackage{soul}
\usepackage{multirow}
\usepackage{graphicx} 
\usepackage{caption}
\usepackage{setspace}

\begin{document}
\begin{frontmatter}
	\title{Manipulating UAV Imagery for Satellite Model Training, Calibration and Testing}
	\author{Jasper Brown$^{a*}$, Cameron Clark$^{b}$, Sabrina Lomax$^{b}$, Khalid Rafique$^{a}$, and Salah Sukkarieh$^{a}$}
	\address{$^{a}$Australian Centre for Field Robotics (ACFR), Faculty of Engineering, The University of Sydney, NSW 2006, Australia}
	\address{$^{b}$Livestock Production and Welfare Group, School of Life and Environmental Sciences, Faculty of Science, The University of Sydney, NSW 2006, Australia}
	\cortext[mycorrespondingauthor]{Corresponding author: j.brown@acfr.usyd.edu.au}
	
	\begin{abstract}
		Modern livestock farming is increasingly data driven and frequently relies on efficient remote sensing to gather data over wide areas. High resolution satellite imagery is one such data source, which is becoming more accessible for farmers as coverage increases and cost falls. Such images can be used to detect and track animals, monitor pasture changes, and understand land use. Many of the data driven models being applied to these tasks require ground truthing at resolutions higher than satellites can provide. Simultaneously, there is a lack of available aerial imagery focused on farmland changes that occur over days or weeks, such as herd movement. With this goal in mind, we present a new multi-temporal dataset of high resolution UAV imagery which is artificially degraded to match satellite data quality. An empirical blurring metric is used to calibrate the degradation process against actual satellite imagery of the area. UAV surveys were flown repeatedly over several weeks, for specific farm locations. This 5cm/pixel data is sufficiently high resolution to accurately ground truth cattle locations, and other factors such as grass cover. From 33 wide area UAV surveys, 1869 patches were extracted and artificially degraded using an accurate satellite optical model to simulate satellite data. Geographic patches from multiple time periods are aligned and presented as sets, providing a multi-temporal dataset that can be used for detecting changes on farms. The geo-referenced images and 27853 manually annotated cattle labels are made publicly available.  
	\end{abstract}
\end{frontmatter}
	

\section{Introduction}
Remote sensing data can support a range of farm functions, from real time automated cattle management, to long term farmer decision support tools. While many of these tools make use of the high resolution imagery provided by unmanned aerial vehicle (UAV) platforms, these are often not scalable to large area cattle stations, and the greater coverage offered by satellite imagery is instead required. Developing and validating precision livestock farming (PLF) techniques for satellite imagery requires accurate ground truth, which is often difficult to match to sparse or unpredictable satellite acquisition dates. In this work we take the opposite approach by gathering high resolution UAV data at short intervals across several weeks, then artificially degrading this to generate simulated satellite imagery which can be used to design or test PLF tools. Surveys are presented as high resolution geo-registered images for ground truthing, along with labelled patch pairs where each element of the pair comes from a different day. Because of this, the dataset can be used for change analysis at high or low resolution, as well as for validating satellite imagery algorithms using the high resolution UAV images.  

To calibrate and validate the degradation approach used, a high resolution satellite image was compared to the UAV data. Spatial resolution, as measured using ground sample distance (GSD), was matched using bicubic downsampling. Optical blurring was applied by calculating the point spread function (PSF) of the simulated camera aperture. Blurring magnitude is measured using the Laplacian kernel variance method and the simulated camera parameters were empirically tuned to match the true satellite imagery kernel variance. This allows us to verify the accuracy of the degradation method. 

Remote sensing using space based, or aerial platforms, can be applied to understand changes in many key farmland metrics. These typically focus on pasture observation and management \citep{ali2016}, or animal movements and behaviours \citep{handcock2009}. High resolution panchromatic satellite images can be used to count large mammals \citep{xue2017}, including cattle \citep{laradji2020}. The lower resolution, lack of ground truth accuracy and high cost of satellite data can limit its applications in agriculture \citep{hollings2018}. Artificially degrading high resolution UAV imagery to match satellite data is one means of avoiding these issues when developing PLF techniques for satellite applications. Our previous work investigated cattle detection using this approach, but lacked real satellite data to empirically validate the method and was not multi-temporal \citep{brown2022}. 

Drone imagery has likewise been used for numerous PLF tasks such as animal detection and behaviour estimation. Many authors have demonstrated cattle and sheep detection using UAV platforms \citep{rivas2018,sarwar2018,barbedo2019,barbedo2020,shao2020}. A large body of literature exists for change detection using remote sensing \citep{pettorelli2005}. However, this is typically only applied to cattle farming settings at long time scales and for wide area changes, such as paddock wide normalised difference vegetation index (NDVI) or land cover type \citep{mas1999}. The dataset presented here can help bridge this gap towards shorter term change analysis at the individual object scale.

To construct this new dataset, drone surveys were repeatedly flown for a series of paddocks, then stitched into a single geo-registered image per flight. These are manually annotated with accurate cattle locations at high resolution. Pairs of patches are extracted from each combination of geo-registered images and aligned using pixel wise phase correlation. These patch pairs are degraded to simulate the ground sample distance of 0.5m/pix satellite imagery. An optical blurring kernel is calculated using simulated camera parameters, which in turn are tuned to match the simulated imagery Laplacian kernel variance to the true satellite imagery variance. Data augmentation is applied to the patch pairs to capture the colouration, misalignment, and stitching artefacts present in the dataset. The original geo-registered images, annotated patch pairs and simulated satellite patch pairs are available from \url{http://data.acfr.usyd.edu.au/Agriculture/RemoteSensingFarmData}.

\section{Method}
To generate the dataset, images were gathered using a quadcopter UAV platform for several locations over a period of two months. Images from each flight are then stitched together into a single top down geo-registered image, similar to a very high resolution satellite image. These geo-images are aligned for each location and split into patches for labelling. Each patch is manually annotated with cattle locations, then is degraded to simulate the resolution and optical quality of commercial satellite imagery. Actual satellite imagery from one of the survey locations is compared to the degraded imagery to validate the degradation method.  

\subsection{UAV Data Gathering}
Images were gathered using a DJI Phantom 3 quadcopter UAV with inbuilt FC300C camera. This has a 20mm focal length with a 6.17x4.55mm CMOS sensor and resolution of 4000x3000 pixels. All surveys were flown at 120m altitude, using a lawnmower pattern generated using the DJI Ground Station Pro flight planning software. Flight paths were planned with a 60\% sideways overlap and 80\% forwards overlap to enable image stitching. Each image is also tagged with its GPS position. 

Images were captured from one site in the Southern Highlands region of Australia and from a site on the outskirts of Sydney. Both sites contain several survey areas, as detailed in Table \ref{tab:flights}. Some images in each flight contain inaccurate GPS data, so are manually removed. The remainder are stitched into a geo-registered image using Agisoft Metashape, resulting in one geo-image per drone flight. 

The Southern Highlands site data targets medium term change analysis, with images gathered daily where the weather permitted. The Sydney data targets short term change analysis, with surveys over 3 consecutive days for the PB and PC sets. The PA1-PA3 set covers a single day, with cattle being manually moved in the paddock between flights. High wind and rain prevented flights during some of the highlands survey periods, so these dates are not evenly distributed. Further details are available in the online dataset.

\begin{table}[]
\resizebox{\textwidth}{!}{%
\begin{tabular}{|c|c|c|c|c|c|}
\hline
\textbf{Designation} & \textbf{Location} & \textbf{Date Range} & \textbf{\begin{tabular}[c]{@{}c@{}}Number \\ of \\ Flights\end{tabular}} & \textbf{\begin{tabular}[c]{@{}c@{}}Number \\ of \\ Images\end{tabular}} & \textbf{\begin{tabular}[c]{@{}c@{}}Number \\ of Patch \\ Pairs\end{tabular}} \\ \hline
A1-A15 & Highlands A & 15/7/2021-17/9/2021 & 15 & 5166 & 2726 \\ \hline
B1-B4 & Highlands B & 20/8/2021-31/8/2021 & 4 & 529 & 52 \\ \hline
C1-C5 & Highlands C & 10/9/2021-17/9/2021 & 5 & 1050 & 162 \\ \hline
PA1-PA3 & Sydney A & 15/7/2021-15/7/2021 & 3 & 384 & 76 \\ \hline
PB1-PB3 & Sydney B & 17/8/2021-19/8/2021 & 3 & 505 & 60 \\ \hline
PC1-PC3 & Sydney C & 7/9/2021-9/9/2021 & 3 & 142 & 12 \\ \hline
\textbf{Total} &  & 15/7/2021-17/9/2021 & 33 & 7776 & 3088 \\ \hline
\end{tabular}%
}
\caption{Details of all UAV flights.} 
\label{tab:flights}
\end{table}

\subsection{Image Patch Generation and Labelling}
Accurate image alignment is important for many difference based change detection approaches, so several alignment steps were applied. The first of these crops the geo-image to the furthest north-west pixel area present in all surveys of a location, such that pixel (0,0) aligns for all images. This means for any of the 15 images in the A series, any given pixel (u,v) will correspond to the same real world location across all images in that series. This requires overlapping coverage for all images in a set.

After cropping, a combined translation, rotation and scaling adjustment is calculated which maximises the pixel phase correlation between the geo-image pairs to be aligned. This was determined using only the centre 2000x2000 pixels from each geo-image to reduce computation requirements. All steps were run on a desktop computer using an Intel Xeon E5-2630 and 32GB of RAM. The geo-images are too large for further efficient processing, so were broken into tiles of 5000x5000 pixels following the alignment step. This was done in such a way that a given patch covers the same geographic area across all flights in that set. For example, a tree at pixel (u,v) in patch 1 of flight A1, will also appear at (u,v) in patch 1 of A2, A3, A4, and so on. Patches are generated in pairs with one from each flight, and all possible combinations of flights are stored as matching patch pairs, so some patches are duplicated.  

\begin{figure*}[h!]
	\centering	
	\includegraphics[width=0.95\textwidth]{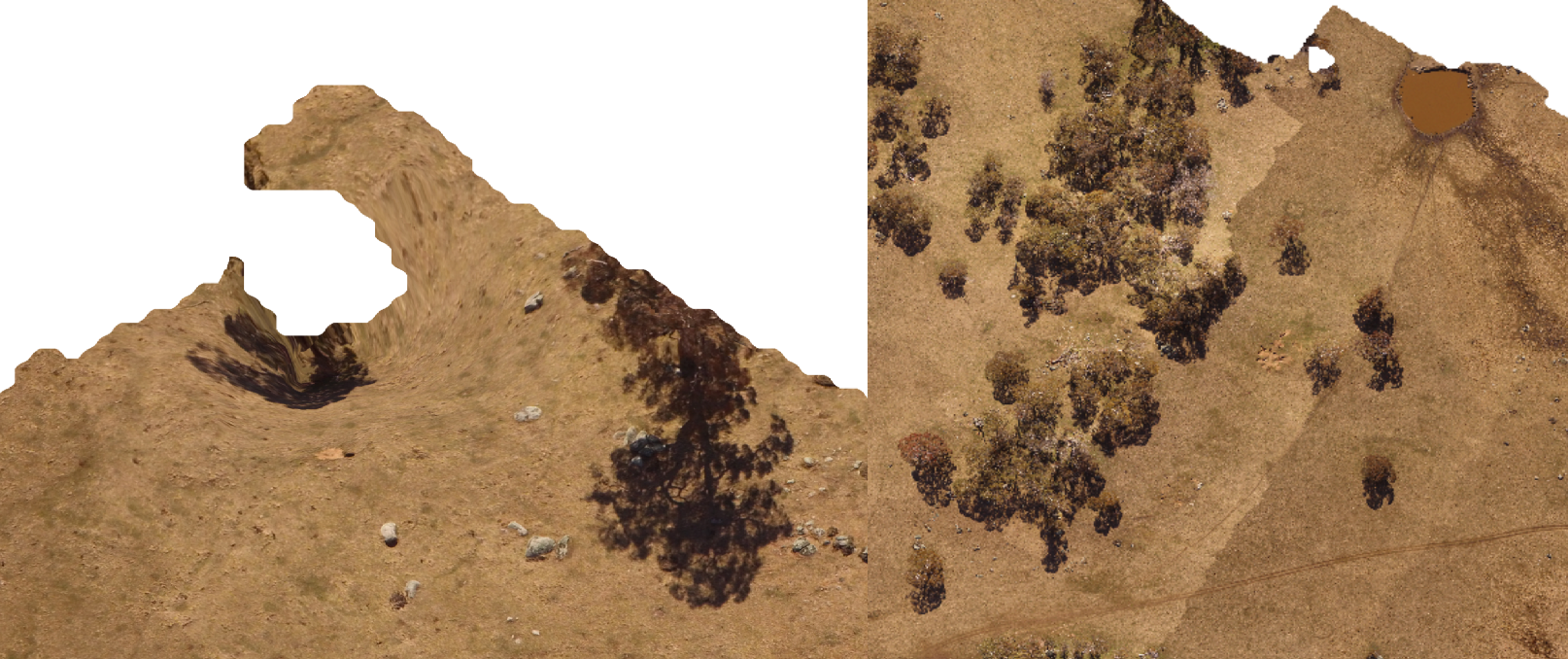}
	\caption{Local warping artefact (left) caused by large tree on the edge of the survey area, diagonal lines of exposure change (right) caused by changing cloud cover during the survey.}
	\label{fig:Artefacts}
\end{figure*}

Some image artefacts were introduced by the stitching process, including local warping and discontinuous changes in image exposure. Two examples of these are shown in Figure \ref{fig:Artefacts}. Patches with obvious warping or burring were manually removed. 

\begin{figure*}[h!]
	\centering	
	\includegraphics[width=0.95\textwidth]{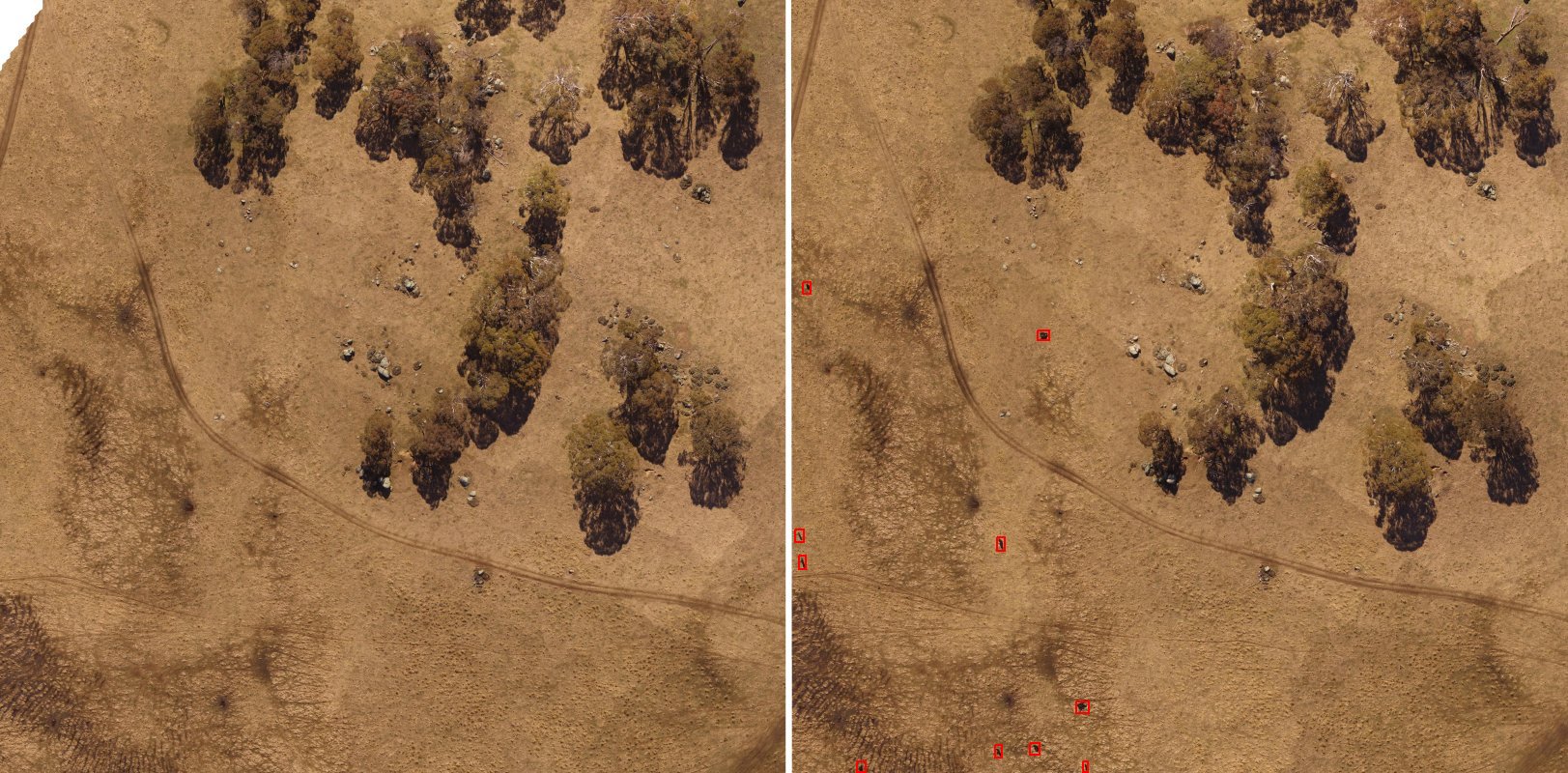}
	\caption{Image patch pair from survey A1 (left) and A2 (right). Cattle can be seen in the lower left quadrant of A2, with annotations visualised as red boxes.}
	\label{fig:Annotated}
\end{figure*}

A focus of this dataset is cattle localisation, so each high resolution patch was labelled with cattle locations and downsampled to simulate satellite imagery. The LabelImg tool from \url{https://github.com/tzutalin/labelImg} was used to manually annotate bounding boxes around each cow using the Yolo label format. Figure \ref{fig:Annotated} shows an example of an annotated patch pair from the A1 and A2 flights. From the 3088 labelled patch pairs, comprising of 1869 unique images, a total of 27853 cattle are annotated.

\subsection{Artificial Degradation}

Once labelled, each patch was degraded using the method from \cite{brown2022} to simulate 50cm satellite imagery. This consists of bicubic downsampling to reach the target ground sample distance and a blurring using a point spread function kernel calculated to match the simulated satellite aperture. The bicubic downsampling ratio is given by the ratio of ground sample distances between the drone and satellite imagery. For all patches, this increases the GSD from 0.05m/pix to 0.5m/pix. 

The simulated aperture function for a hypothetical satellite camera is calculated to match the required Q value. The fast Fourier transform of this is then used to construct the PSF. A circular aperture function is used in this work, and Q is defined as

\begin{equation}
    Q = 2 \frac{resolution_{optical}}{resolution_{detector}} = \frac{\lambda f}{D p}
\end{equation}

where $\lambda$ is the light wavelength, is $f$ focal length, $D$ is the aperture diameter and $p$ is the sensor pixel pitch. The PSF is then applied by convolving it with each pixel, giving a full degradation operation of 

\begin{equation}
I_{sat} = (I_{uav} \circledast PSF)\downarrow _\phi
\label{eqn:degradation}
\end{equation}

where $I_{uav}$ is the drone image patch, $I_{sat}$ is the simulated satellite image, $\circledast$ is the convolution operator and $\downarrow _\phi$ is the bicubic downsampling operation with a GSD ratio of $\phi$.

\subsection{Degradation Calibration and Validation}
Accurately simulating satellite data requires matching the GSD and optical quality of the true satellite data. This captures the key spatial resolution changes, though imaging factors such as white balance, saturation and chromatic aberration are not considered in the present work. Ground sample distance is easily measured and adjusted using downsampling, but optical blurring is more complex and has many contributing factors. So GSD is directly set to match the satellite resolution of 50cm/pix and Q is tuned to minimise the difference between the amount of blurring measured in the degraded UAV imagery, and actual satellite images of the same location. 

Many methods exist for measuring image blur, one common technique which offers a good balance of simplicity and robustness is the Laplacian kernel variance \citep{pertuz2013}. This is calculated by convolving the image with a kernel that approximates the Laplacian mask, then calculating the variance of the resulting image to yield a scalar blur metric \citep{pech2000}. The kernel used is 

\begin{equation}
    \mathcal{L} = \begin{pmatrix}
0.17 & 0.67 & 0.17 \\
0.67 & -3.33 & 0.67 \\
0.17 & 0.67 & 0.17
\end{pmatrix}
\end{equation}

giving a Laplacian Variance (LV) blurring metric of 

\begin{equation}
L = \mathcal{L} \circledast I 
\end{equation}

\begin{equation}
LV = \frac{1}{N} \sum_{u}^{U} \sum_{v}^{V} (L_{u,v} - \Bar{L})^2
\end{equation}

for a single grey-scale input image $I$ where $U,V$ are the input image size, $N=UV$ and $\Bar{L}$ is the mean of all the pixels in $L$. 

\subsection{Data Augmentation}
\begin{figure*}[h!]
	\centering	
	\includegraphics[width=0.95\textwidth]{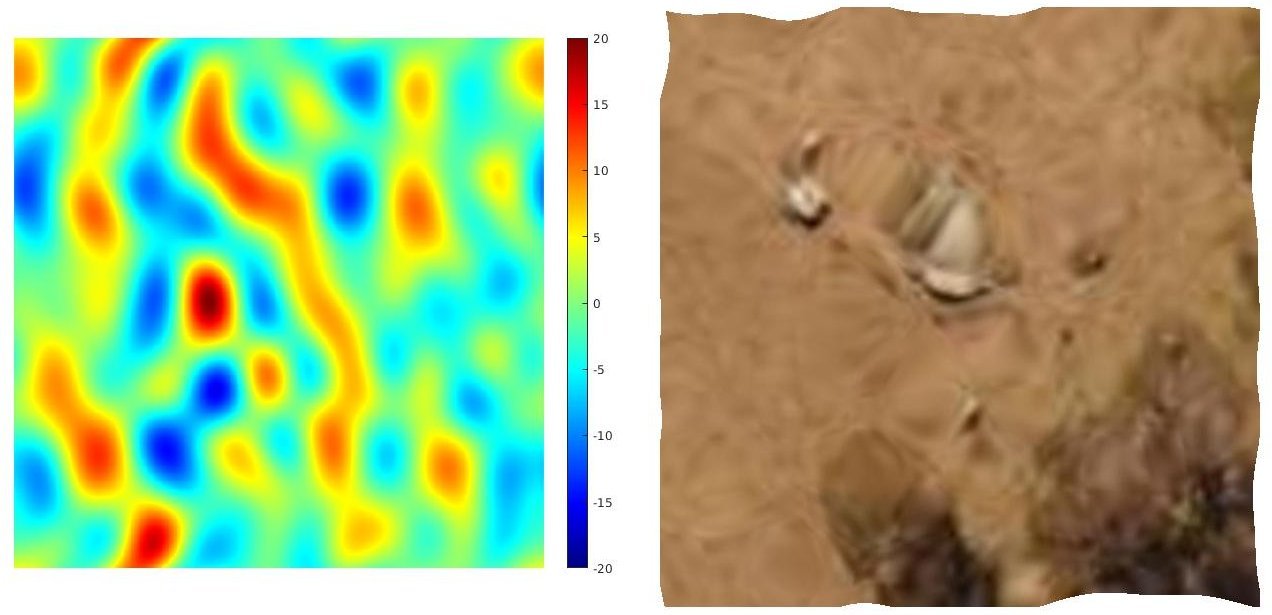}
	\caption{An example of the local warping effect with the u-pixel component mapping surface (left) with units of pixel offset, and a resulting warped image from the downsampled satellite data (right). Warping amount is exaggerated here for clarity.}
	\label{fig:Warping}
\end{figure*}

Data augmentation is a common technique for expanding training dataset size and variety when applying learning-based algorithms. Because of the aligned-pairs nature of the data, most off the shelf augmentation libraries will not work properly with this dataset. So, a series of data augmentation functions are also provided, as listed in Table \ref{tab:augmentation}. The local warping augmentation attempts to capture some of the warping artefacts introduced by the image stitching process. This constructs a random smooth vector field by applying a frequency space filter to random noise. The vector field maximum value and filter width are input parameters. Image pixels are translated according to the vector field with bicubic interpolation used to fill gaps. Figure \ref{fig:Warping} shows an example field and resulting image. 

\begin{table}[]
\resizebox{\textwidth}{!}{%
\begin{tabular}{|c|c|c|}
\hline
\textbf{Function} & \textbf{Description} & \textbf{Parameters} \\ \hline
Rotate & \begin{tabular}[c]{@{}c@{}}Rotate pair by amount sampled \\ uniformly from {[}min,max{]}\end{tabular} & \begin{tabular}[c]{@{}c@{}}min, most negative \\ rotation amount\\ max, most positive \\ rotation amount\end{tabular} \\ \hline
Alignment & \begin{tabular}[c]{@{}c@{}}Shift only one of the image \\ pair slightly to create misalignment\end{tabular} & \begin{tabular}[c]{@{}c@{}}shift standard deviation, \\ where shift is pixel amount \\ sampled from a zero mean Gaussian\end{tabular} \\ \hline
Shift & Shift pair of images together & \begin{tabular}[c]{@{}c@{}}shift standard deviation, \\ where shift is pixel amount \\ sampled from a zero mean Gaussian\end{tabular} \\ \hline
Colour & \begin{tabular}[c]{@{}c@{}}Shift each channel in RGB \\ colour space to recolour image\end{tabular} & \begin{tabular}[c]{@{}c@{}}shift standard deviation, \\ shift is {[}-255,255{]} and sampled \\ from a zero mean Gaussian\end{tabular} \\ \hline
Hue & \begin{tabular}[c]{@{}c@{}}Shift the image hue in HSV \\ colour space to recolour the image\end{tabular} & \begin{tabular}[c]{@{}c@{}}shift standard deviation, \\ shift is {[}-255,255{]} and sampled \\ from a zero mean Gaussian\end{tabular} \\ \hline
Saturation & \begin{tabular}[c]{@{}c@{}}Shift the image saturation \\ in HSV colour space\end{tabular} & \begin{tabular}[c]{@{}c@{}}shift standard deviation, \\ shift is {[}-255,255{]} and sampled \\ from a zero mean Gaussian\end{tabular} \\ \hline
Value & \begin{tabular}[c]{@{}c@{}}Shift the image value in \\ HSV colour space\end{tabular} & \begin{tabular}[c]{@{}c@{}}shift standard deviation, \\ shift is {[}-255,255{]} and sampled \\ from a zero mean Gaussian\end{tabular} \\ \hline
Mirror & \begin{tabular}[c]{@{}c@{}}Flip pair up/down and left/right. \\ Chance of each flip type is \\ sampled independently.\end{tabular} & \begin{tabular}[c]{@{}c@{}}probability, modeled as one Bernoulli \\ distribution for up/down and \\ another for left/right\end{tabular} \\ \hline
Scale & Scale pair by a random amount. & \begin{tabular}[c]{@{}c@{}}scale ratio standard deviation, \\ sampled from a zero mean Gaussian\end{tabular} \\ \hline
Noise & \begin{tabular}[c]{@{}c@{}}Add independent random noise \\ to each image pixel\end{tabular} & \begin{tabular}[c]{@{}c@{}}noise standard deviation, \\ noise is {[}-255,255{]}x3 and \\ sampled from a zero mean Gaussian\end{tabular} \\ \hline
Shear & \begin{tabular}[c]{@{}c@{}}Apply a shear geometric \\ operation to pair\end{tabular} & \begin{tabular}[c]{@{}c@{}}shear standard deviation, \\ shear is a ratio sampled \\ from a zero mean Gaussian\end{tabular} \\ \hline
Local Warp & See Above & \begin{tabular}[c]{@{}c@{}}max surface height and \\ warp together or independently\end{tabular} \\ \hline
\end{tabular}%
}
\caption{The augmentation functions provided, their effect on the image patch pair and the tuning parameters for each.}
\label{tab:augmentation}
\end{table}

\section{Degradation Results}
\begin{figure*}[h!]
	\centering	
	\includegraphics[width=0.95\textwidth]{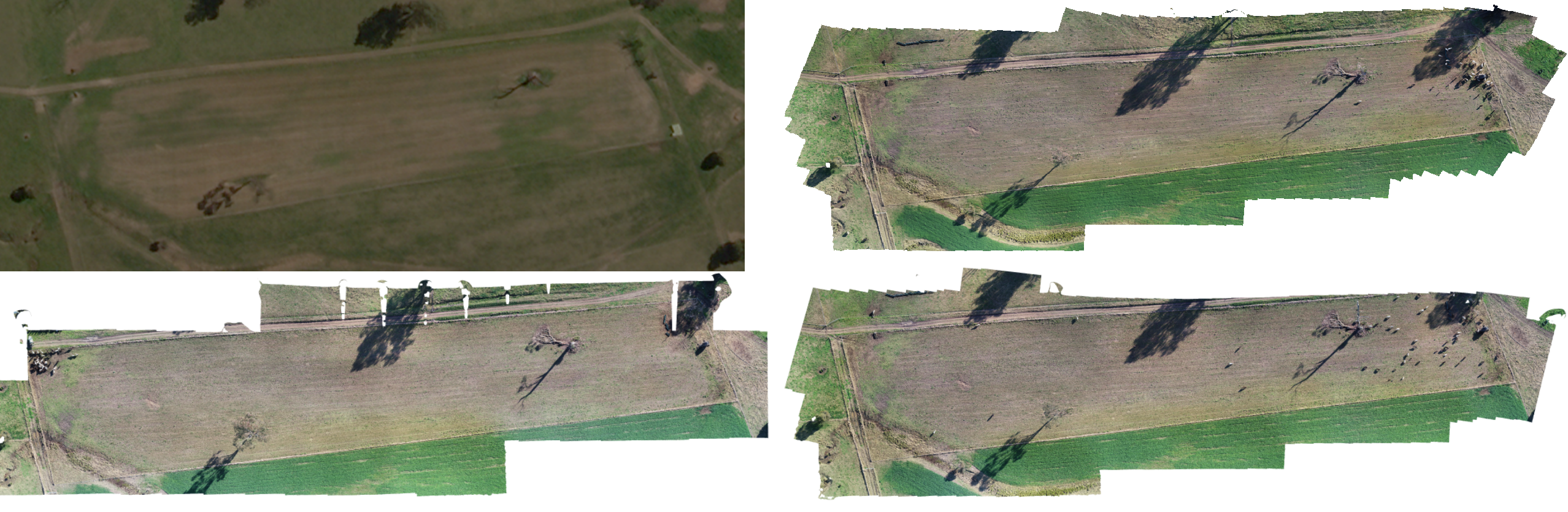}
	\caption{Comparison of the (clockwise from top left) true satellite image patch, and high resolution UAV surveys PA1, PA2, PA3.}
	\label{fig:satcomparison}
\end{figure*}

\begin{figure*}[h!]
	\centering	
	\includegraphics[width=0.95\textwidth]{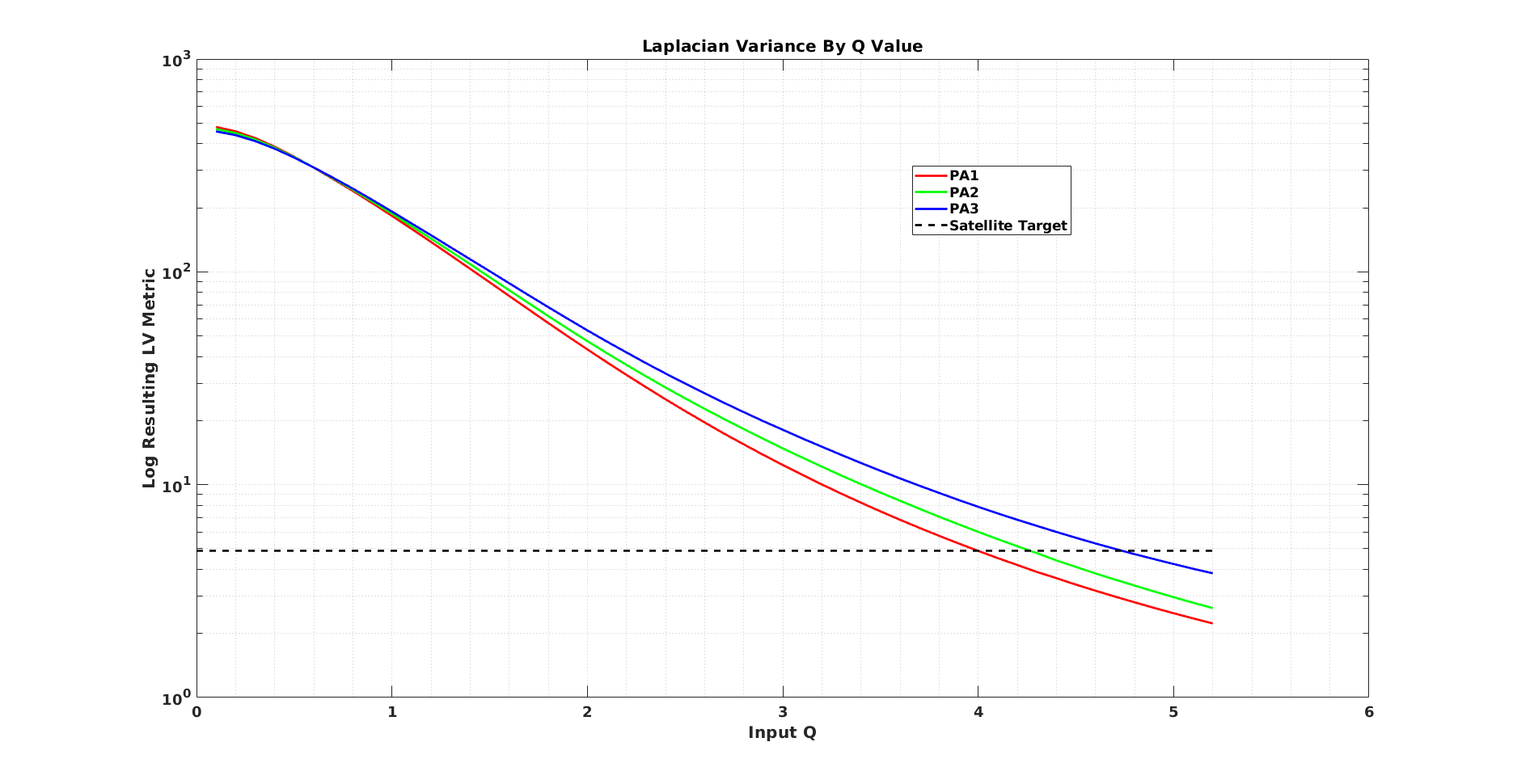}
	\caption{Laplacian variance blurring metric for the degraded UAV image by chosen Q value for the 3 PA survey flights.}
	\label{fig:qsweep}
\end{figure*}

To calibrate the degradation optical blurring step, an image patch is extracted from commercial satellite imagery covering the PA survey areas. A Planet SkySat Collect image is used for calibration, this is 0.5m/pix GSD with an off nadir angle of 25.3\textdegree ~\citep{planet2017}. Only the RGB spectral bands are used. This is cropped to the same extents as the three UAV geo-registered images, shown in Figure \ref{fig:satcomparison}. An LV metric of 4.89 is calculated for the satellite patch, and a sweep over Q values is run to determine the Q value which yields an equivalent LV metric for the degraded UAV imagery. An average Q of 4.34 is calculated for the three surveys, as shown in Figure \ref{fig:qsweep}.

\begin{figure*}[h!]
	\centering	
	\includegraphics[width=0.95\textwidth]{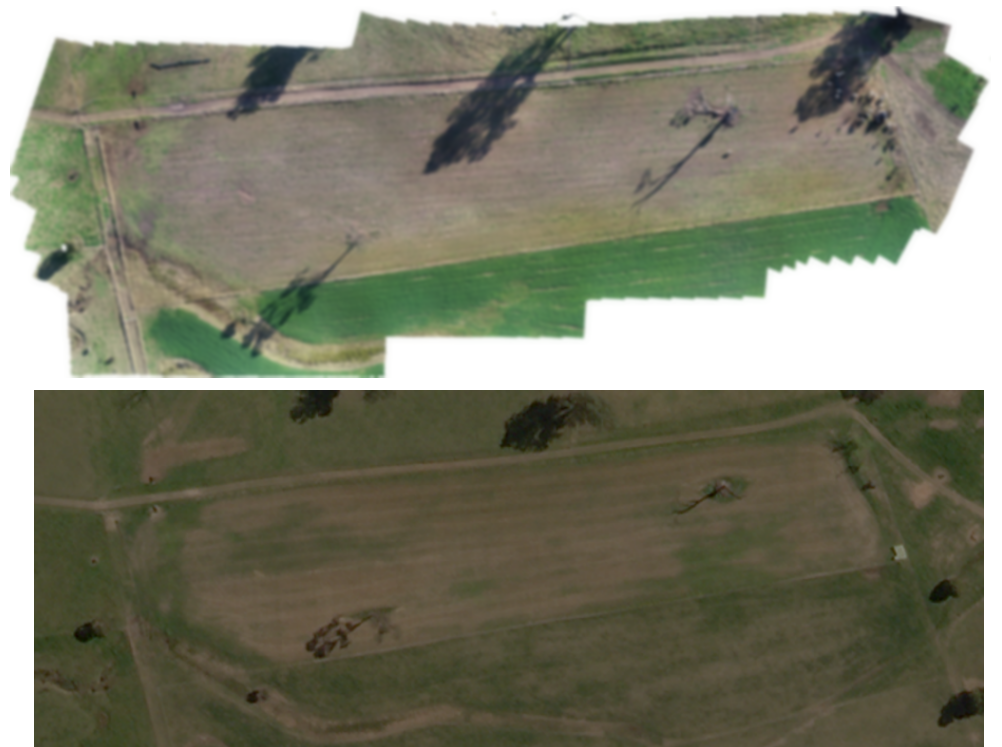}
	\caption{The degraded UAV image using Q of 4.34 and GSD of 0.5m/pix (top), and the original satellite image (bottom). No colour adjustment has been performed.}
	\label{fig:comparison}
\end{figure*}

\begin{figure*}[h!]
	\centering	
	\includegraphics[width=0.95\textwidth]{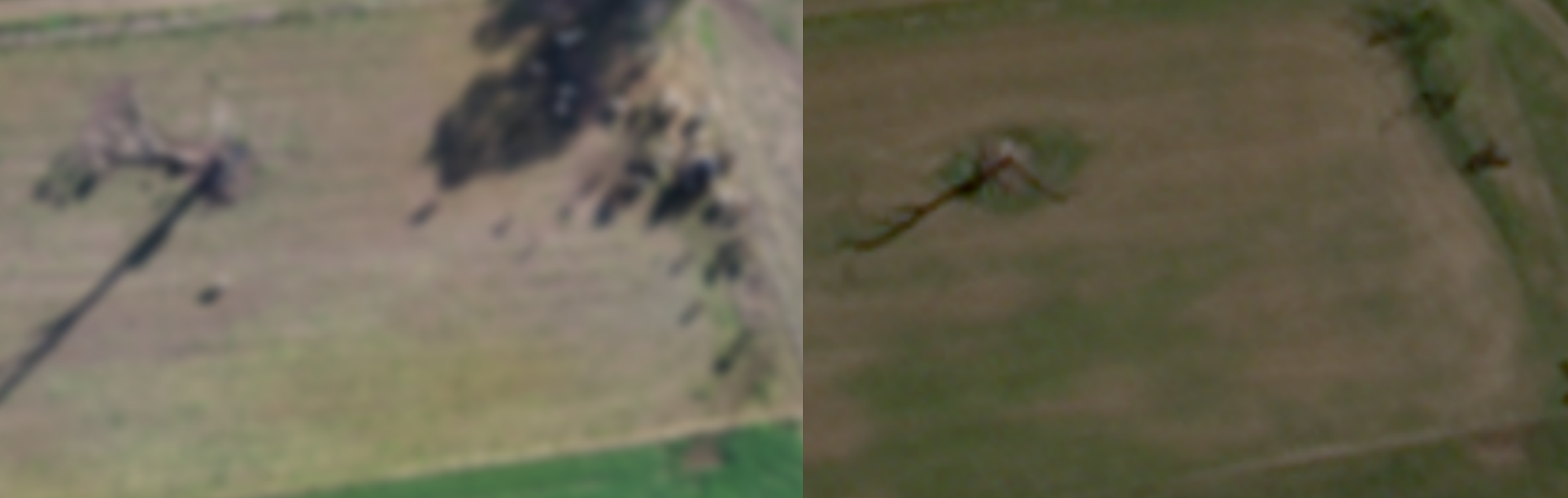}
	\caption{Zoomed view of a region of the degraded UAV image (left) and the true satellite image (right).}
	\label{fig:zoomed}
\end{figure*}

After applying this level of blurring, the simulated and actual satellite images can be compared in Figure \ref{fig:comparison}. Our method does not attempt to correct for differences in colouring, brightness or shadow. The satellite capture is from September 21st 2021 collected at 9:56am local time, while the PA surveys occurred on July 17th 2021 around mid afternoon. Figure \ref{fig:zoomed} shows a zoomed view of the same data. 

The average Q value of 4.34 required to match the satellite LV is higher than typical camera values of 0.5 to 2.0. This is likely driven by additional blurring factors beyond optical diffraction, such as atmospheric and motion blur, which are empirically captured in the Q value. Blurring metrics are also sensitive to sensor type, so assessing the suitability of the Laplacian variance measure to agricultural satellite imagery would be beneficial future work. 

\section{Conclusion}
Degrading high resolution UAV imagery to simulate satellite data was shown to be a viable approach to developing satellite remote sensing datasets while maintaining accurate ground truth. Though the high Q value estimated from the calibration step indicates that empirical calibration using at least one satellite image of the surveyed area has significant benefits. Methods for accurately aligning and augmenting multi-temporal UAV surveys were also presented. 

The labelled dataset, as both patch pairs and geo-images, can be accessed at \url{http://data.acfr.usyd.edu.au/Agriculture/RemoteSensingFarmData}. Data augmentation code and further details of survey dates and areas are also available there. We anticipate this dataset will have applications for cattle detection across multiple time scales, as well as short term change detection for farm environments. Both the high resolution drone images and their simulated satellite counterparts are available, so high resolution ground truth can be used to assess satellite imagery farm analysis tools. 

\section*{ACKNOWLEDGMENTS}
The authors acknowledge the support of the Meat \& Livestock Australia Donor Company through the project: Objective, robust, real-time animal welfare measures for the Australian red meat industry (P.PSH.0819). We thank Paul Lipscombe for his on-farm assistance with data gathering and Planet for providing satellite data. 

\section*{CRediT authorship contribution statement}
J.B.: Conceptualization, Investigation, Data curation, Methodology, Validation, Formal analysis, and Writing-original draft; C.C: Funding acquisition, Writing - review \& editing; S.L: Writing - review \& editing; K.R: Resources, Writing - review \& editing; S.S.: Resources, Writing - review \& editing.

\section*{Declaration of competing interest}
The authors declare that they have no known competing financial interests or personal relationships that could have appeared to influence the work reported in this paper.

\newpage
\bibliography{ManipulatingDroneImagery}
\bibliographystyle{apalike}

\end{document}